\documentclass[letterpaper, 10 pt, conference]{ieeeconf}
\IEEEoverridecommandlockouts
\usepackage{xurl}
\usepackage[utf8]{inputenc}
\usepackage[T1]{fontenc}
\usepackage{graphicx}
\usepackage{booktabs}
\usepackage{threeparttable}
\usepackage{graphics}
\usepackage{epsfig}
\usepackage{mathptmx}
\usepackage{mathptmx}
\usepackage{amsmath}
\usepackage{amssymb}
\usepackage{xcolor}
\usepackage{multirow}
\usepackage{multicol}
\usepackage{algorithmic}
\usepackage{algorithm}

\title{\LARGE \bf Uncertainty-Aware Trip Purpose Inference from GPS Trajectories via POI Semantic Zones and Pareto Calibration}

\author{
	\parbox{\textwidth}{
		\centering
		Bo Yang$^{1,*}$, Haoxuan Ma$^{1,*,\dagger}$, Yifan Liu$^{1}$, Zhiyuan Zhang$^{1}$, Chris Stanford$^{2}$, Morgan Sun$^{2}$, and Jiaqi Ma$^{1}$
	}
	\thanks{$^{*}$Equal contribution}
    \thanks{$^{\dagger}$Corresponding Author: haoxuanma@ucla.edu}
	\thanks{$^{1}$UCLA Mobility Lab, Department of Civil and Environmental Engineering, University of California, Los Angeles, Los Angeles, USA.}
	\thanks{$^{2}$Novateur Research Solutions, Ashburn, VA, USA.}
}

\begin{document}

\maketitle
\thispagestyle{empty}
\pagestyle{empty}

\begin{abstract}
	Large-scale GPS trajectory data offer rich observations of human mobility,
	yet assigning trip purposes to detected stops remains challenging due to
	the absence of individual-level ground truth, spatial uncertainty from GPS
	noise and incomplete points of interest (POIs) coverage, and fundamental
	behavioral differences across trip purposes.
	We propose a weakly supervised framework integrating neighborhood-level POI
	semantic zones with distance-weighted spatial likelihoods, differentiated
	inference strategies for mandatory and non-mandatory activities, and a
	multi-phase Pareto optimization that jointly minimizes distributional
	divergence from household travel survey statistics and maximizes inference
	reliability without requiring annotated labels.
	Evaluated on over 81 million staypoints in Los Angeles, the framework reduces activity type frequency Jensen-Shannon distance (JSD) by 23\%, start time JSD by 48\%, and duration JSD by 12\% respectively relative to a
	comparable baseline. The proposed approach provides a scalable and
	uncertainty-aware path from raw GPS trajectories to semantically annotated
	mobility data for travel demand modeling and transportation policy analysis.
\end{abstract}

\section{Introduction}
\label{sec:introduction}

\begin{figure*}[!htbp]
	\centering
	\includegraphics[width=\textwidth]{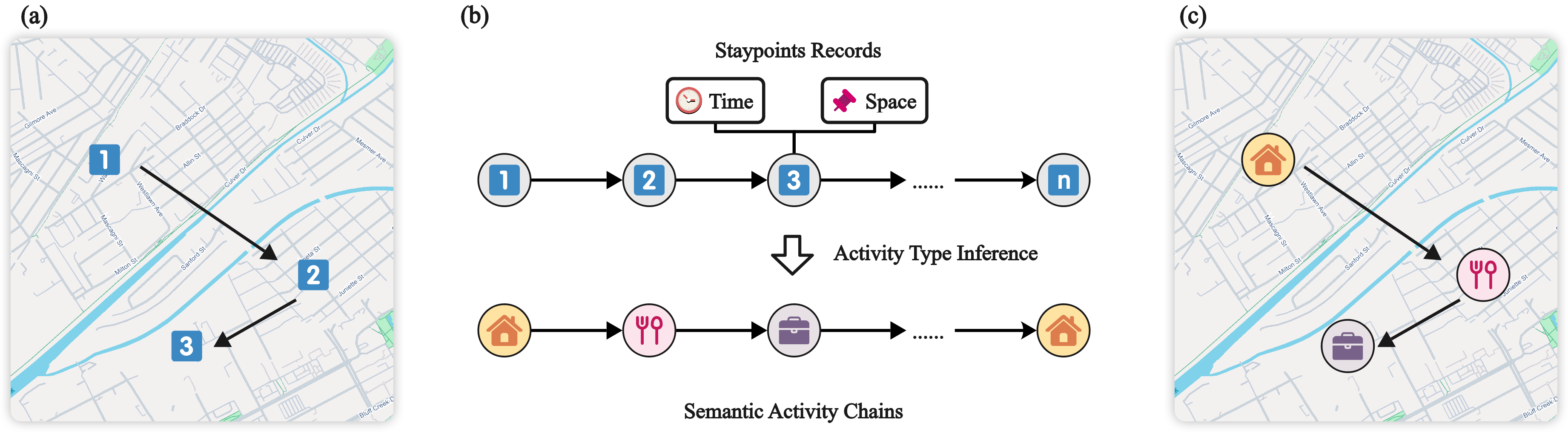}
	\caption{Illustration of the activity type inference task. (a) Raw staypoints extracted from GPS trajectories, where spatiotemporal stop episodes are detected but carry no semantic label. (b) Activity type inference assigns an activity category to each staypoint (e.g., Home, Work, Dining) by reasoning over spatial context and temporal patterns. (c) The resulting semantic activity chain represents the individual's daily mobility as an ordered sequence of labeled stops, enabling behaviorally meaningful analysis.}
	\label{fig:problem}
\end{figure*}

With the widespread adoption of smartphones and passive sensing systems, large-scale GPS trajectory datasets have grown rapidly, enabling research on traffic demand forecasting \cite{barbosa2018human, necula2014dynamic}, travel behavior analysis \cite{kraemer2020mapping}, and next-generation activity-based demand modeling \cite{ma2025learning}.
Raw trajectories are commonly segmented into sequences of staypoints, where stationary episodes are detected when trajectory points remain within a spatial threshold for a minimum duration \cite{zheng2015trajectory}. This segmentation results in discrete episodes with identifiable spatial locations and temporal time frames. However, staypoints capture only the spatiotemporal geometry; the behavioral purpose of each stop remains latent. In travel demand analysis, trip purpose is conventionally defined by the activity conducted at the trip destination. Therefore, inferring trip purpose from GPS trajectories can be operationalized as identifying the activity type associated with each detected staypoint (such as Home, Work, or Shopping). Recovering these activity types provides the semantic foundation for interpreting mobility behavior, constructing human mobility generation models, and ensuring consistency with survey-based statistics \cite{petrillo2024editorial, liao2025next, 10920138}.
Therefore, we focus on trip purpose inference by assigning activity types to detected staypoints using trajectory-derived and contextual information (Figure~\ref{fig:problem}).

Activity type inference from GPS trajectories remains an open challenge, as the problem is simultaneously label-scarce, spatially uncertain, and behaviorally heterogeneous: ground-truth activity labels at the staypoint level are rarely available at scale \cite{parent2013semantic}, and existing approaches do not provide an explicit measure of inference reliability \cite{bohte2009deriving, deng2010deriving}, limiting the ability to systematically validate and calibrate models in the absence of individual-level annotations; validating against travel surveys is further complicated by structural mismatches between GPS-observed and survey-reported patterns, where shift worker commutes and irregular travel tend to be underreported in household surveys \cite{stopher2007assessing, bricka2012analysis}. Spatial semantics introduce another layer of difficulty, as GPS noise propagates through staypoint detection and complicates location-to-activity association \cite{gong2015identification, huang2019transport}; contextual sources such as points of interest (POIs) from OpenStreetMap (OSM) \cite{OpenStreetMap2017} are often incomplete or unevenly distributed \cite{psyllidis2022points}, rendering deterministic nearest-POI matching brittle and prone to systematic bias.
Beyond spatial uncertainty, activity behavior is intrinsically heterogeneous: mandatory activities exhibit stable spatial anchors and strong temporal regularity, whereas non-mandatory activities are episodic and context-dependent \cite{schlich2003habitual}, necessitating differentiated inference strategies rather than uniform rules.

To address these challenges, we propose a framework for activity type inference that operates under limited supervision, differentiates inference strategies by activity type, and models spatial semantics probabilistically. The main contributions of this paper are as follows:

\begin{itemize}
	\item We propose a machine-learning-integrated weakly supervised framework that calibrates inference parameters by aligning inferred activity distributions with NHTS \cite{nrel_tsdc_2019} statistics without staypoint-level ground truth, assigns explicit confidence scores to track per-staypoint inference reliability, and jointly optimizes distributional alignment and reliability via multi-phase Pareto optimization.
	\item We develop differentiated inference strategies reflecting that mandatory activities exhibit stable spatial anchors and temporal regularity, motivating cross-day spatiotemporal aggregation, while non-mandatory activities are episodic and context-dependent, requiring probabilistic integration of spatial context, temporal priors, and duration with bias correction for GPS--survey discrepancies.
	\item We introduce neighborhood-level POI semantic zones that aggregate distance-weighted POI semantics into spatially coherent activity representations, replacing brittle nearest-POI matching and improving robustness to GPS positional error and uneven POI coverage.
\end{itemize}

\section{Literature Review}
\label{sec:literaturereview}

\subsection{Regularity-Driven and Heuristic Inference}
\label{sec:lit-rulebased}

Early approaches to activity type inference exploit recurring spatiotemporal patterns through handcrafted rules based on visit frequency, time-of-day windows, and spatial stability \cite{zhao2020discovering, stopher2008deducing}, which are well suited to mandatory activities that exhibit stable anchors and strong behavioral regularity. Subsequent work extends this foundation by incorporating probabilistic elements, combining POI proximity, temporal likelihoods, and survey-derived priors to increase flexibility while preserving interpretability \cite{yin2021exploring, liu2021semantic, zahedi2018estimating}. However, the underlying logic remains anchored to recurrent behavior, and performance degrades for non-mandatory activities that are episodic, spatially dispersed, and context-dependent. Reliance on fixed temporal heuristics and local POI matching introduces sensitivity to GPS noise, incomplete POI coverage, and mixed-use environments \cite{amini2022combined}.

\subsection{Supervised and Data-Driven Classification}
\label{sec:lit-supervised}

Supervised approaches frame activity inference as a classification problem over trajectory-derived and contextual features. Traditional machine learning models \cite{lu2019random, bolbol2012inferring}, sequential architectures such as Long Short-Term Memory networks \cite{chen2020towards}, and graph-based models \cite{lyu2021graph, corrias2023exploring, shi2021sgcn, fan2024global} can capture richer spatial and temporal dependencies than rule-based systems and achieve strong predictive performance when labeled data are available.
Nevertheless, staypoint-level annotations are costly to obtain, prone to labeling error, and difficult to generalize across cities and populations \cite{parent2013semantic}. More fundamentally, they optimize for individual-level prediction accuracy without constraining aggregate activity distributions \cite{yin2017generative}, and the absence of explicit confidence estimation limits applicability where inference reliability must be assessed without ground truth.

\subsection{Semantic Enrichment with POI Metadata and LLMs}
\label{sec:lit-poi-llm}

Recent work increasingly leverages POI metadata to enrich spatial semantics, with large language models (LLMs) emerging as a flexible means of mapping heterogeneous POI descriptions to activity categories without task-specific supervision \cite{10920138, wang2024large, zhang2024agentic}. These methods broaden semantic coverage and reduce dependence on manually defined POI taxonomies. Despite this progress, most implementations treat POIs as isolated points and assign activity labels through nearest-POI matching or independent semantic scoring \cite{long2025harnessing}, remaining sensitive to GPS positional error, spatial misalignment, and uneven POI distributions \cite{psyllidis2022points}. In dense or mixed-use areas, a single nearest POI is often insufficient to represent the surrounding functional context. GPS noise further compounds this difficulty, as positional errors introduced during trajectory collection and staypoint detection propagate into subsequent semantic association steps \cite{ibrahim2021gps, hwang2018segmenting, zhu2023difftraj}, making robust spatial representation a prerequisite for reliable inference. Duration information is similarly underexploited: survey-derived duration distributions are commonly applied directly, without accounting for systematic discrepancies between consolidated survey episodes and the typically shorter GPS staypoints \cite{stopher2007assessing}.

\section{Methodology}
\label{sec:methodology}

\begin{figure*}[t]
	\centering
	\includegraphics[width=\textwidth]{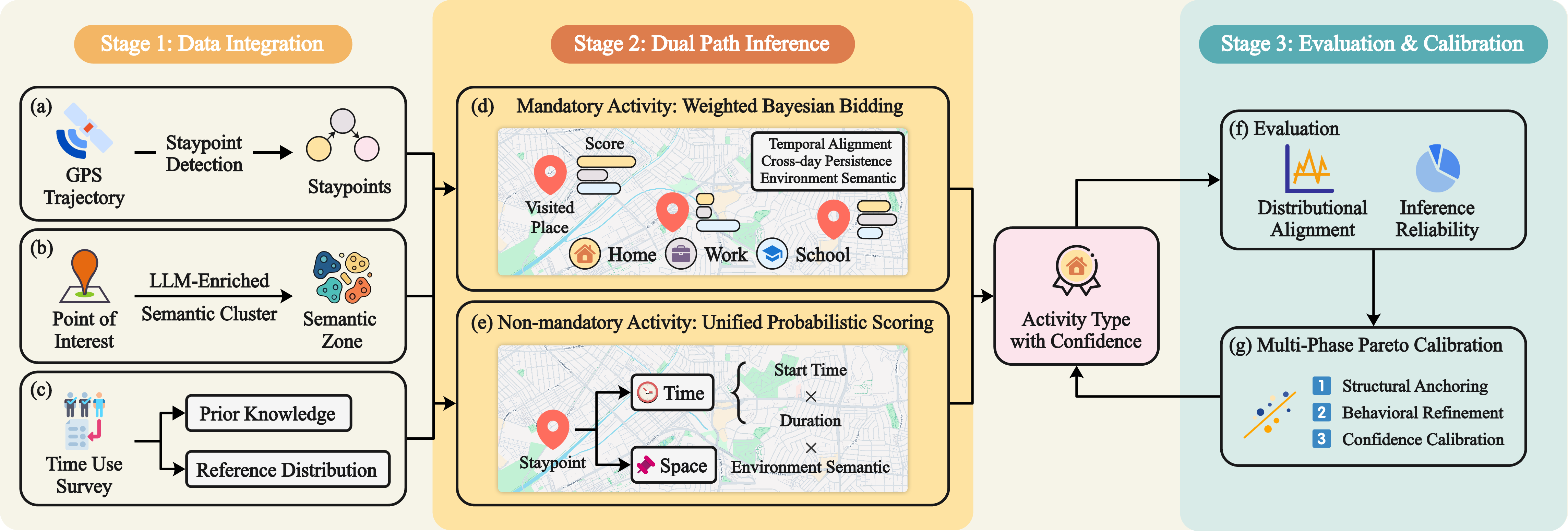}
	\caption{Three-stage activity inference pipeline. Stage 1 (a) extracts staypoints from raw trajectory data, (b) constructs semantic zones from POIs, and (c) derives reference distributions from NHTS. Stage 2 infers an activity type and confidence score for each staypoint: (d) mandatory activities are identified by exploiting spatial recurrence and temporal regularity on frequently visited locations, whereas (e) non-mandatory activities are assigned by integrating spatial context, temporal priors, and activity duration. Stage 3 (f) evaluates inference quality using distributional alignment and inference reliability, and (g) optimizes framework parameters via multi-phase Pareto optimization with these two measures as joint objectives.}
	\label{fig:pipeline}
\end{figure*}

\subsection{Problem Formulation}
\label{subsec:problem_formulation}

We formulate GPS-based activity inference as a weakly supervised probabilistic reasoning problem under behavioral heterogeneity and uncertain spatial semantics. The framework outputs activity labels and calibrated confidence scores, and is optimized through distribution alignment with survey-derived reference statistics rather than staypoint-level supervision. Figure~\ref{fig:pipeline} illustrates the overall three-stage pipeline.

\subsubsection{Data Representations}

\textbf{Staypoints.}
Let $\mathcal{U}=\{u_1,\dots,u_N\}$ denote a population of agents.
For each agent $u$, raw GPS trajectories are segmented into staypoints
$\mathcal{S}_u=\{s_1,\dots,s_{K_u}\}$.
Each staypoint is represented as
\begin{equation}
	s_i=\langle l_i, t_{\mathrm{start},i}, t_{\mathrm{end},i}, d_i\rangle,
\end{equation}
where $l_i = (\mathrm{lat}_i, \mathrm{lon}_i)$ denotes the geographic location, $t_{\mathrm{start},i}$ and $t_{\mathrm{end},i}$ denote arrival and departure times, and $d_i = t_{\mathrm{end},i} - t_{\mathrm{start},i}$ denotes stay duration.

\textbf{Activity Space.}
Let $\mathcal{A}$ denote the activity set, partitioned into mandatory activities
$\mathcal{A}_{\mathrm{mand}}=\{\mathrm{Home},\mathrm{Work},\mathrm{School}\}$ and non-mandatory activities
$\mathcal{A}_{\mathrm{non}}=\mathcal{A}\setminus\mathcal{A}_{\mathrm{mand}}$, reflecting distinct behavioral regularities and motivating differentiated inference mechanisms described in Sections~\ref{subsec:mandatory_inference} and~\ref{subsec:nonmandatory_inference}.

\textbf{Semantic Context.}
Spatial semantics are provided by a POI set $\mathcal{P}$, where each $p_j \in \mathcal{P}$ is associated with geographic coordinates and metadata including names and descriptions.

\textbf{Reference Knowledge.}
Household survey statistics $\mathcal{K}$ provide reference distributions over activity type frequency, start times, and durations, serving as structural constraints for parameter calibration without staypoint-level ground truth.

\subsubsection{Inference Objective}

The framework learns a mapping
\begin{equation}
	(\hat{y}_i,\hat{c}_i)=f_{\theta}(s_i,\mathcal{P},\mathcal{K}),
\end{equation}
where $\hat{y}_i\in\mathcal{A}$ is the inferred activity label and $\hat{c}_i\in[0,1]$ is a confidence score reflecting inference reliability.

Because staypoint-level labels are unavailable, parameter calibration relies on distributional alignment. Let $\mathcal{D}_{\mathrm{infer}}(\theta)$ denote the inferred distributions over activity type frequency, start time, and duration aggregated across the population, and let $\mathcal{D}_{\mathrm{ref}}$ denote the corresponding survey-derived reference distributions. Parameters are selected to satisfy:
\begin{equation}
	\min_{\theta} \mathrm{JSD}\bigl(\mathcal{D}_{\mathrm{infer}}(\theta),\mathcal{D}_{\mathrm{ref}}\bigr),
	\qquad
	\max_{\theta} \mathrm{HCR}(\theta),
\end{equation}
where JSD denotes Jensen--Shannon Distance \cite{luca2021survey} and High-Confidence Ratio (HCR) is defined as the fraction of staypoints whose confidence score meets or exceeds a threshold, reflecting the proportion of inferences deemed reliable. As these two objectives may conflict, they are handled jointly via Pareto optimization, described in Section~\ref{subsec:parameter_optimization}.

\subsection{Semantic Environment Representation}
\label{subsec:semantic_environment}

GPS positional error, uneven POI coverage, and dense mixed-use environments mean that a single nearest POI often fails to represent the functional character of a location reliably. We therefore construct a neighborhood-level semantic representation by aggregating POIs into spatial zones, and derive spatial likelihoods through distance-weighted zone scoring rather than point-level matching.

\subsubsection{POI Semantic Enrichment}

Each POI $p_j \in \mathcal{P}$ is assigned a probability distribution over activity types, where each entry represents the likelihood that a visit to $p_j$ serves a given activity purpose: $\mathbf{v}_j=[P(a_1|p_j),\dots,P(a_{|\mathcal{A}|}|p_j)]$.
This distribution is obtained via LLM-based semantic inference \cite{10920138}, which maps heterogeneous POI metadata to activity-level probabilities without requiring manually defined taxonomies.

\subsubsection{Semantic Zone Construction and Spatial Likelihood}

Enriched POIs are spatially clustered using DBSCAN \cite{ester1996density} to form semantic zones, with isolated points assigned to singleton zones to ensure full spatial coverage. Each zone $c$ is characterized by its centroid $l_c$, radius $\rho_c$, and a normalized aggregate distribution $\mathbf{v}_c = \mathrm{normalize}(\sum_{j \in c} \mathbf{v}_j)$ over activity types. For any query location $l$ (e.g.\ a staypoint or a cluster centroid), the spatial likelihood for activity type $a$ is
\begin{equation}
	\label{eq:spatial_likelihood}
	P_{\mathrm{space}}(a \mid l) \propto
	\sum_{c\in\mathcal{N}(l)}
	\exp\!\left(-\frac{d_{lc}^2}{2\sigma^2}\right)\mathbf{v}_c(a),
\end{equation}
where $\mathcal{N}(l)$ is the set of zones within a search radius of $l$, $d_{lc} = \mathrm{dist}(l, l_c)$, and $\sigma$ is a bandwidth controlling spatial decay. This soft, distance-weighted aggregation reduces sensitivity to individual missing or misplaced POIs and yields a spatially smooth semantic representation.

\subsection{Mandatory Activity Inference: Weighted Bayesian Bidding}
\label{subsec:mandatory_inference}

Mandatory activities (Home, Work, School) exhibit strong spatial recurrence and temporal regularity, motivating an inference strategy that aggregates evidence across multiple observations of the same location over time. For agent $u$, staypoints are spatially clustered, and clusters visited on at least two distinct days are retained as candidate locations $\mathcal{L}_u=\{\ell_1,\dots,\ell_M\}$. Each cluster is represented by a duration-weighted centroid, and staypoints are snapped to these centroids to reduce GPS noise and spatial dispersion.

Each candidate location $\ell$ competes for each mandatory activity
$a \in \mathcal{A}_{\mathrm{mand}}$ by accumulating time-use evidence
across visits and days, obtained by integrating the survey-derived
time-of-day prior $P_{\mathrm{ref}}(a, t)$ over the stay interval and
aggregating across days with anomaly suppression to reduce the influence
of atypical visits. The bid score is defined as
\begin{equation}
	B(\ell,a)
	=
	P_{\mathrm{time}}(a \mid \ell)
	\cdot
	w_{\mathrm{stab}}(\ell)
	\cdot
	P_{\mathrm{space}}(a \mid \ell),
\end{equation}
where the temporal term $P_{\mathrm{time}}(a \mid \ell)$ aggregates time-use evidence over all visits to $\ell$, with daily cap $\tau_a$: $P_{\mathrm{time}}(a \mid \ell) = \sum_{d \in \mathrm{days}(\ell)} \min\bigl( \sum_{s \in \mathcal{S}(\ell,d)} \int_{[s]} P_{\mathrm{ref}}(a,t)\,\mathrm{d}t, \tau_a \bigr)$, with $[s]$ the time interval of stay $s$ and $\mathcal{S}(\ell,d)$ the set of staypoints at $\ell$ on day $d$.
$w_{\mathrm{stab}}(\ell) = \log_2(1 + |\mathrm{days}(\ell)|)$ captures cross-day persistence of the location, and
$P_{\mathrm{space}}(a \mid \ell)$ provides soft semantic prior from the surrounding POI environment.
The multiplicative combination integrates temporal, stability, and spatial evidence into a unified bidding score.

For each activity, the highest-scoring candidate is selected only if it
shows sufficient dominance, preventing mandatory assignment when evidence
is inadequate. Confidence is defined as the normalized margin between the
top two bid scores, quantifying how decisively the selected location exceeds
its closest competitor, suited to mandatory inference where uncertainty lies
in distinguishing among a small set of recurrent locations rather than a
broad activity-type space. Algorithm~\ref{alg:mandatory_bidding} summarizes
the procedure.
\begin{algorithm}[ht]
	\caption{Mandatory activity inference via weighted Bayesian bidding}
	\label{alg:mandatory_bidding}
	\small
	\begin{algorithmic}[1]
		\STATE \textbf{Input:} candidate locations $\mathcal{L}_u$;
		staypoints $\mathcal{S}_u$;
		prior $P_{\mathrm{ref}}(a,t)$;
		spatial likelihood $P_{\mathrm{space}}(a \mid \ell)$;
		daily cap $\tau_a$;
		existence threshold $\theta_{\mathrm{exist}}$
		\STATE \textbf{Output:} $\ell_{\mathrm{home}}$, $\hat{c}_{\mathrm{home}}$;
		$\ell_{\mathrm{mand}}$, $a$, $\hat{c}_{\mathrm{mand}}$
		\FOR{each $\ell \in \mathcal{L}_u$, $a \in \mathcal{A}_{\mathrm{mand}}$}
		\FOR{each day $d \in \mathrm{days}(\ell)$}
		\STATE $\mathrm{daily}(\ell,a,d) \leftarrow
			\min\!\bigl(\sum_{s \in \mathcal{S}(\ell,d)}
			\int_{[s]} P_{\mathrm{ref}}(a,t)\,\mathrm{d}t,\;
			\tau_a\bigr)$
		\ENDFOR
		\STATE $P_{\mathrm{time}}(a \mid \ell) \leftarrow
			\sum_{d} \mathrm{daily}(\ell,a,d)$
		\quad $\triangleright$ zero on weekends for Work/School
		\STATE $w_{\mathrm{stab}}(\ell) \leftarrow
			\log_2(1 + |\mathrm{days}(\ell)|)$
		\STATE $B(\ell,a) \leftarrow
			P_{\mathrm{time}}(a \mid \ell) \cdot
			w_{\mathrm{stab}}(\ell) \cdot
			P_{\mathrm{space}}(a \mid \ell)$
		\ENDFOR
		\STATE $\ell_{\mathrm{home}} \leftarrow \arg\max_{\ell}\,
			B(\ell, \mathrm{Home})$
		\STATE $\ell_2 \leftarrow \arg\max_{\ell \in \mathcal{L}_u \setminus
				\{\ell_{\mathrm{home}}\}} B(\ell, \mathrm{Home})$
		\STATE $\hat{c}_{\mathrm{home}} \leftarrow
			\mathrm{clamp}\!\left(
			\tfrac{B(\ell_{\mathrm{home}},\mathrm{Home}) -
				B(\ell_2,\mathrm{Home})}{B(\ell_{\mathrm{home}},\mathrm{Home})},
			0, 1\right)$
		\STATE $\mathcal{L}' \leftarrow
			\mathcal{L}_u \setminus \{\ell_{\mathrm{home}}\}$
		\STATE $\ell_{\mathrm{mand}} \leftarrow
			\arg\max_{\ell \in \mathcal{L}'}\, B(\ell, \mathrm{Work})$
		\IF{$B(\ell_{\mathrm{mand}}, \mathrm{Work}) < \theta_{\mathrm{exist}}$}
		\RETURN $\ell_{\mathrm{home}}$, $\hat{c}_{\mathrm{home}}$,
		no Work/School
		\ENDIF
		\STATE $a \leftarrow \arg\max_{a' \in \{\mathrm{Work,School}\}}
			P_{\mathrm{space}}(a' \mid \ell_{\mathrm{mand}})$
		\STATE $\ell_2' \leftarrow \arg\max_{\ell \in \mathcal{L}'
				\setminus \{\ell_{\mathrm{mand}}\}} B(\ell, a)$
		\STATE $\hat{c}_{\mathrm{mand}} \leftarrow
			\mathrm{clamp}\!\left(
			\tfrac{B(\ell_{\mathrm{mand}},a) -
				B(\ell_2',a)}{B(\ell_{\mathrm{mand}},a)},
			0, 1\right)$
		\RETURN $\ell_{\mathrm{home}}$, $\hat{c}_{\mathrm{home}}$,
		$\ell_{\mathrm{mand}}$, $a$, $\hat{c}_{\mathrm{mand}}$
	\end{algorithmic}
\end{algorithm}

\subsection{Non-Mandatory Activity Inference: Unified Probabilistic Scoring}
\label{subsec:nonmandatory_inference}

Non-mandatory activities exhibit substantially higher variability in timing,
location choice, and dwell duration, rendering hard decision rules unstable.
Accordingly, we adopt a unified probabilistic scoring framework that integrates
spatial, temporal, and duration evidence in a unified multiplicative form.

For each staypoint $s_i$, the score for activity type
$k \in \mathcal{A}_{\mathrm{non}}$ is defined as
\begin{equation}
	S_k
	=
	\bigl(P_{\mathrm{space}}(k) + \epsilon\bigr)
	\bigl(P_{\mathrm{tod}}(t_i \mid k) + \epsilon\bigr)
	\bigl(P_{\mathrm{dur}}(b_i \mid k) + \epsilon\bigr)^{\alpha},
\end{equation}
where $P_{\mathrm{space}}(k)$ is the spatial likelihood from \eqref{eq:spatial_likelihood} at $l_i$,
$P_{\mathrm{tod}}(t_i \mid k)$ is the start time prior $P(\text{start} = t_i \mid k)$ from survey statistics $\mathcal{K}$,
and $P_{\mathrm{dur}}(b_i \mid k)$ is the duration prior for the bin $b_i$ of stay duration $d_i$, from $\mathcal{K}$.
The floor $\epsilon > 0$ ensures numerical stability when any component is zero.
The exponent $\alpha$ regulates the contribution of duration evidence.

A key difficulty arises from the structural mismatch between survey-based activity records and GPS-derived staypoints: survey durations are discretized and represent consolidated episodes, whereas GPS observations contain a large number of brief stays, causing direct use of survey duration distributions to over-penalize short visits. To account for this discrepancy, duration evidence is incorporated as soft behavioral support through two complementary mechanisms. The exponent $\alpha$ controls the degree to which duration influences the final score, downweighting its contribution for short-duration bins where GPS--survey discrepancies are most pronounced; this parameter is subsequently calibrated via the Pareto optimization framework described in Section~\ref{subsec:parameter_optimization}.
A duration prior correction further adjusts duration distribution shape by reallocating limited probability mass toward shorter bins for activity types commonly associated with brief engagements, while preserving the overall preferences ordering.

The final activity label is assigned as $\hat{y}_i = \arg\max_k S_k$.
Confidence is defined as $\hat{c}_i = \max_k \mathrm{posterior}(k)$, where $\mathrm{posterior}(k) = S_k / \sum_{k'} S_{k'}$.
Since non-mandatory scores are derived from a unified multiplicative framework over all $k \in \mathcal{A}_{\mathrm{non}}$, the normalized posterior naturally quantifies how decisively one activity dominates the full competing set, which a two-way margin would only partially reflect. Algorithm~\ref{alg:nonmandatory_scoring} summarizes the procedure.
\begin{algorithm}[ht]
	\caption{Non-mandatory activity inference via unified probabilistic scoring}
	\label{alg:nonmandatory_scoring}
	\small
	\begin{algorithmic}[1]
		\STATE \textbf{Input:} staypoint $s_i$ $(l_i, t_i, d_i)$;
		spatial likelihood $P_{\mathrm{space}}(k)$;
		priors $P_{\mathrm{tod}}(t \mid k)$, $P_{\mathrm{dur}}(b \mid k)$
		from $\mathcal{K}$; floor $\epsilon$; duration-bin-dependent exponent $\alpha$
		\STATE \textbf{Output:} $\hat{y}_i \in \mathcal{A}_{\mathrm{non}}$,
		$\hat{c}_i \in [0,1]$
		\STATE $t \leftarrow \mathrm{slot}(t_i)$;
		$b_i \leftarrow \mathrm{bin}(d_i)$
		\FOR{each $k \in \mathcal{A}_{\mathrm{non}}$}
		\STATE $S_k \leftarrow
			\bigl(P_{\mathrm{space}}(k)+\epsilon\bigr)
			\bigl(P_{\mathrm{tod}}(t \mid k)+\epsilon\bigr)
			\bigl(P_{\mathrm{dur}}(b_i \mid k)+\epsilon\bigr)^{\alpha}$
		\ENDFOR
		\STATE $Z \leftarrow \sum_{k} S_k$
		\IF{$Z \leq 0$ \textbf{or} $Z$ is not finite}
		\STATE $\mathrm{posterior}(k) \leftarrow
			1/|\mathcal{A}_{\mathrm{non}}|$
		for all $k$ \quad $\triangleright$ fallback uniform
		\ELSE
		\STATE $\mathrm{posterior}(k) \leftarrow S_k / Z$
		for all $k$
		\ENDIF
		\STATE $\hat{y}_i \leftarrow \arg\max_{k}\, S_k$;
		$\hat{c}_i \leftarrow \max_k\, \mathrm{posterior}(k)$
		\RETURN $\hat{y}_i$, $\hat{c}_i$
	\end{algorithmic}
\end{algorithm}

\subsection{Parameter Optimization: Multi-Phase Pareto Framework}
\label{subsec:parameter_optimization}

Calibrating the framework requires simultaneously balancing distributional fidelity and inference reliability across numerous interacting parameters, where a single scalar objective often yields unstable solutions.
To overcome this, we adopt a multi-phase Pareto optimization strategy based on NSGA-II \cite{deb2002fast}, progressively calibrating parameter subsets while keeping previously optimized components fixed. In the first phase (structural anchoring), global parameters controlling spatial influence and aggregation are tuned to minimize JSD over activity type and start time distributions. In the second phase (behavioral refinement), parameters governing the integration of spatial, temporal, and duration evidence for non-mandatory activities are adjusted to improve start time and duration alignment, without altering the global activity structure established in the first phase. In the third phase (confidence calibration), parameters controlling confidence scores are refined to maximize HCR while keeping activity assignments fixed, ensuring that confidence values reflect inference robustness rather than posterior sharpness. This decoupled strategy produces stable and interpretable parameter settings under weak supervision.

\section{Experiment}
\label{sec:experiment}

\subsection{Experimental Setup}
\label{subsec:experimental_setup}

\subsubsection{Dataset}
\label{subsec:dataset}

Experiments are conducted in the Los Angeles metropolitan area using a
large-scale anonymized mobile-device GPS trajectory dataset provided by
Veraset \cite{veraset2024}, covering the period from January to May 2019.
After staypoint extraction and filtering for agents with sufficient
observation records, the dataset comprises 23,601,957 daily trajectories and 81,731,392 staypoints from
426,875 agents. Spatial semantic context is derived from OSM POI data
\cite{OpenStreetMap2017} for the same region, comprising 434,305 POIs. Activity reference distributions are derived from LA-specific records
extracted from the NHTS \cite{nrel_tsdc_2019} to ensure geographic
representativeness of the reference statistics used for evaluation.

\subsubsection{Activity Taxonomy}

The activity type set $\mathcal{A}$ follows the NHTS taxonomy with $|\mathcal{A}|=15$ types, partitioned into three mandatory activities ($\mathcal{A}_{\mathrm{mand}}$) and twelve non-mandatory activities ($\mathcal{A}_{\mathrm{non}}$), as listed in Table~\ref{tab:activity_taxonomy}.

\begin{table}[htbp]
	\centering
	\caption{Activity type taxonomy ($|\mathcal{A}| = 15$).}
	\label{tab:activity_taxonomy}
	\begin{tabular}{c|l|c|l|c|l}
		\hline
		\multicolumn{6}{l}{\textbf{Mandatory Activities}}      \\
		\hline
		1  & Home       & 2  & Work       & 3  & School        \\
		\hline
		\multicolumn{6}{l}{\textbf{Non-Mandatory Activities}}  \\
		\hline
		4  & Caregiving & 5  & Shop Goods & 6  & Shop Services \\
		7  & Meals Out  & 8  & Errands    & 9  & Leisure       \\
		10 & Exercise   & 11 & Social     & 12 & Healthcare    \\
		13 & Worship    & 14 & Other      & 15 & Pickup/Drop   \\
		\hline
	\end{tabular}
\end{table}

\subsubsection{Evaluation Metrics}

Given the absence of staypoint-level ground truth, evaluation is conducted by comparing inferred activity distributions against NHTS reference distributions using Jensen--Shannon Distance, a symmetric and bounded divergence measure defined as:
\begin{equation}
	\mathrm{JSD}(P \| Q) = \frac{1}{2}KL(P \| M) + \frac{1}{2}KL(Q \| M),
\end{equation}
where $M = \frac{1}{2}(P + Q)$ and $KL(P \| Q) = \sum_i P(i) \log \frac{P(i)}{Q(i)}$. JSD takes values in $[0,1]$, with lower values indicating better alignment with the reference.

We report three JSD components: activity type frequency JSD, start time JSD, and duration JSD. Start times and durations are discretized into 15-minute bins to match NHTS resolution.
Temporal JSD is computed per activity type and aggregated using NHTS activity shares as weights, ensuring proportional contribution by participation rate and thereby preventing rare activities from dominating the metric.
We additionally report the High-Confidence Ratio with threshold $\tau_c = 0.5$, defined as the proportion of staypoints with predicted confidence $c_{i} \geq \tau_c$. While JSD evaluates population-level distributional alignment, HCR captures instance-level reliability.

\subsubsection{Baseline}

The proposed framework is compared against the activity inference method of \cite{10920138}, which infers activity types from large-scale GPS data by combining spatial semantics and temporal priors with survey-informed constraints. Its probabilistic structure and use of behavioral reference statistics make it a structurally appropriate benchmark for evaluating distributional alignment under weak supervision.

\subsection{Performance Analysis}
\label{subsec:performance}

\subsubsection{Comparison with Baseline}

Table~\ref{tab:baseline_comparison} reports JSD results for the proposed framework and the baseline method. The proposed method achieves lower divergence across all three evaluated dimensions, indicating consistent improvement in both categorical and temporal alignment with the NHTS reference. The most substantial gain is observed in start time JSD, which decreases from 0.5594 to 0.2883, a relative reduction of approximately 48\%. Activity type frequency JSD decreases from 0.2455 to 0.1893, and duration JSD decreases from 0.4864 to 0.4264. The comparatively smaller improvement in duration JSD reflects the inherent difficulty of reconciling GPS-derived stay durations with survey-reported consolidated episodes, as discussed in Section~\ref{subsec:nonmandatory_inference}.

\begin{table}[htbp]
	\centering
	\caption{JSD comparison with baseline.}
	\label{tab:baseline_comparison}
	\begin{tabular}{lcc}
		\hline
		\textbf{Metric}             & \textbf{Baseline} & \textbf{Proposed} \\
		\hline
		Activity Type Frequency JSD & 0.2455            & 0.1893            \\
		Start Time JSD              & 0.5594            & 0.2883            \\
		Duration JSD                & 0.4864            & 0.4264            \\
		\hline
	\end{tabular}
\end{table}

\begin{figure*}[t]
	\centering
	\includegraphics[width=\textwidth]{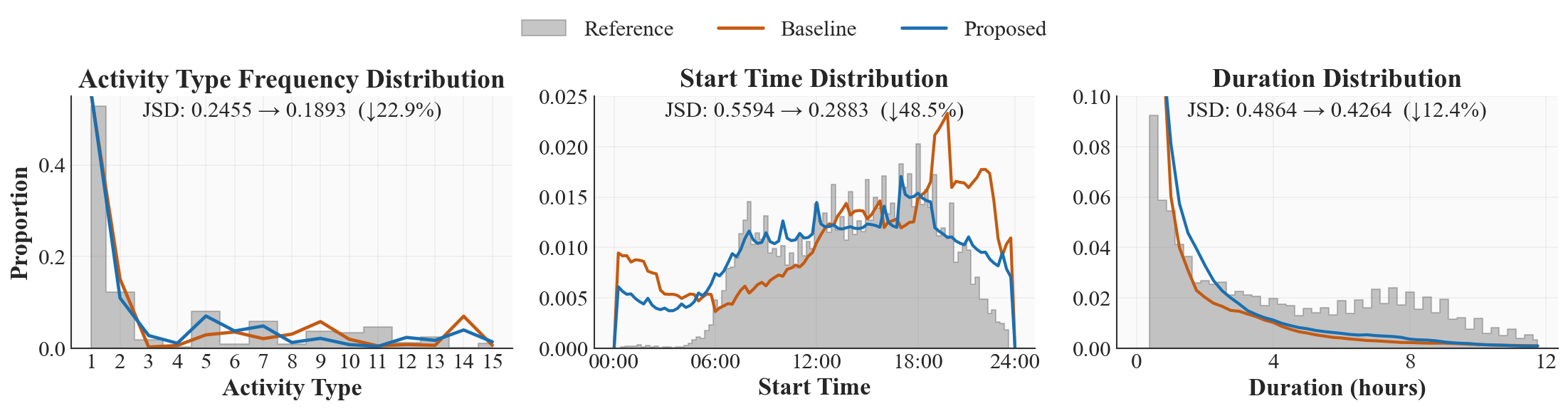}
	\caption{Distributional alignment comparison across activity type frequency, start time, and duration distributions.}
	\label{fig:baseline_comparison}
\end{figure*}

Figure~\ref{fig:baseline_comparison} presents the full empirical distributions for all three components. The proposed method exhibits visibly closer alignment with the reference distributions, most prominently in the temporal dimension. The framework successfully recovers the morning (8:00–10:00) and afternoon (17:00–19:00) activity peaks, and the substantial reduction in start time divergence indicates improved recovery of daily activity rhythms, suggesting that the differentiated inference strategies better capture the temporal structure of each activity class. Gains in activity type frequency and duration distributions further demonstrate enhanced preservation of structural composition and temporal allocation patterns relative to the baseline.

A residual divergence persists in the
early morning period (before 6:00), where GPS trajectories capture activity
initiations underrepresented in the NHTS reference. This gap is
consistent with two known sources of GPS--survey structural mismatch:
shift worker trips beginning outside conventional commute windows
\cite{ma2025learning, ma2025beyond} and emergency or irregular trips
respondents tend to omit or consolidate in household travel surveys
\cite{stopher2007assessing, bricka2012analysis}. This divergence reflects a structural gap between data sources;
the framework's preservation of these off-peak GPS-observed patterns while
maintaining strong alignment during mainstream activity hours suggests
the proposed approach balances fidelity to survey reference distributions
with the behavioral richness captured in GPS trajectories.

\subsubsection{Effects of Multi-Phase Pareto Calibration}

Table~\ref{tab:phase_effects} reports the evolution of all JSD components and HCR across the three calibration phases, starting from the uncalibrated initial parameter configuration.

\begin{table}[htbp]
	\centering
	\caption{JSD components and HCR across calibration phases. \\ M = mandatory; N = non-mandatory.}
	\label{tab:phase_effects}
	\resizebox{\columnwidth}{!}{%
		\scriptsize
		\begin{tabular}{lcccc@{\hskip 6pt}c}
			\hline
			\textbf{Phase} & \textbf{Freq.} & \textbf{Start} & \textbf{Dur.} & \textbf{HCR (M)} & \textbf{HCR (N)} \\
			\hline
			Original       & 0.2018         & 0.4278         & 0.4550        & 82.7\%           & 38.7\%           \\
			Phase~1        & 0.1907         & 0.2910         & 0.4312        & 82.6\%           & 40.3\%           \\
			Phase~2        & 0.1894         & 0.2883         & 0.4285        & 82.6\%           & 41.7\%           \\
			Phase~3        & 0.1893         & 0.2883         & 0.4264        & 83.1\%           & 41.7\%           \\
			\hline
		\end{tabular}%
	}
\end{table}

Phase 1 (structural anchoring) produces the largest single-phase improvement:
start time JSD decreases from 0.4278 to 0.2910 and activity type JSD from
0.2018 to 0.1907. Duration JSD also improves from 0.4550 to 0.4312 despite
not being a direct objective, suggesting that structural activity alignment
indirectly benefits temporal allocation patterns. Phase 2 (behavioral refinement) further reduces duration JSD from 0.4312 to
0.4285 and raises non-mandatory HCR from 40.3\% to 41.7\%, while previously
calibrated components remain stable or marginally improve, confirming that the
staged design prevents regression of earlier objectives. Phase 3 (confidence calibration) refines confidence scores without modifying
activity assignments, increasing mandatory HCR from 82.6\% to 83.1\% while
all JSD metrics remain stable. Overall, all metrics improve monotonically or
remain stable across phases, validating the staged Pareto calibration design.

\subsubsection{Confidence Validity under Positional Perturbation}

Gaussian noise was added to trajectory coordinates at three levels ($\sigma =
	5$, 10, and 20 meters) to simulate realistic GPS positional error. Perturbed
staypoints were matched to the original dataset using agent ID, start time,
and end time; match rates exceeded 99.5\% across all noise levels, demonstrating
that our staypoint extraction procedure is robust to positional noise
at the evaluated levels. Stability is reported as the proportion of matched records
with unchanged inferred activity type, stratified by confidence level
(high: $\hat{c}_i \ge 0.5$; low: $\hat{c}_i < 0.3$).

\begin{figure}[htbp]
	\centering
	\includegraphics[width=\columnwidth]{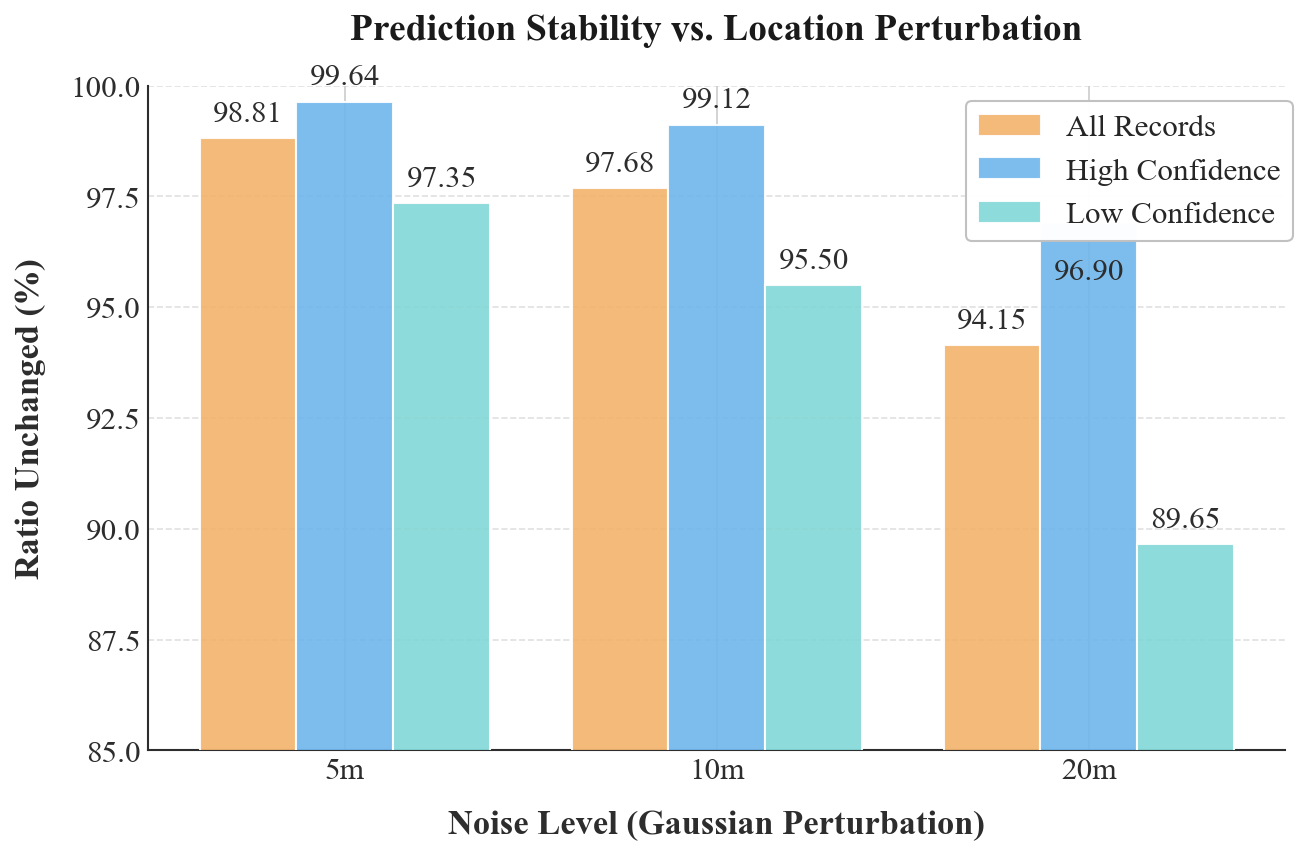}
	\caption{Prediction stability across confidence levels under Gaussian
		location perturbation ($\sigma = 5$, 10, 20 meters).}
	\label{fig:noise_stability}
\end{figure}

Figure~\ref{fig:noise_stability} and Table~\ref{tab:noise_per_activity}
present the results. The framework achieves weighted average stability of
98.81\%, 97.68\%, and 94.15\% at 5, 10, and 20 meters respectively,
demonstrating robustness across all noise levels.
High-confidence records consistently outperform low-confidence records at
every level: the stability gap increases from 2.29 percentage points at
5 meters (99.64\% vs.\ 97.35\%) to 7.25 percentage points at 20 meters
(96.90\% vs.\ 89.65\%), confirming the confidence score becomes
increasingly discriminative as positional error grows.

Table~\ref{tab:noise_per_activity} further shows that mandatory activities
(Home: 99.99\%, Work: 99.22\%, School: 98.06\%) achieve substantially higher
stability than non-mandatory activities, with the lowest stability observed
for Shop Services (91.19\%) and Leisure (92.27\%). This reflects the
underlying inference mechanism: mandatory activities are identified through
cross-day location recurrence and are inherently insensitive to small
positional displacements, whereas non-mandatory inference relies on
distance-weighted POI zone scoring, where noise shifts the effective
neighborhood and alters spatial likelihoods.

\begin{table}[!htbp]
	\centering
	\caption{Per-activity prediction stability under 10m GPS perturbation.}
	\label{tab:noise_per_activity}
	\begin{tabular}{cc|cc}
		\hline
		\textbf{Activity type} & \textbf{Ratio (\%)} & \textbf{Activity type}          & \textbf{Ratio (\%)} \\
		\hline
		1                      & 99.99               & 9                               & 92.27               \\
		2                      & 99.22               & 10                              & 94.38               \\
		3                      & 98.06               & 11                              & 97.04               \\
		4                      & 95.48               & 12                              & 92.43               \\
		5                      & 93.06               & 13                              & 93.45               \\
		6                      & 91.19               & 14                              & 93.19               \\
		7                      & 92.83               & 15                              & 94.72               \\
		8                      & 92.24               & \textbf{Activity-Weighted Avg.} & \textbf{97.68}      \\
		\hline
	\end{tabular}
\end{table}

\subsubsection{Confidence Validity under POI Incompleteness}

While the previous experiment evaluates positional error sensitivity,
this experiment examines a complementary dimension: robustness to incomplete
semantic context. POIs were randomly deleted at 5\% and 10\% rates, and
stability was stratified by confidence level using the same
protocol.

The framework achieves a weighted average stability of 92.43\% under 10\%
POI deletion, reflecting the importance of spatial semantic context to
inference quality. High-confidence records again maintain substantially higher
stability than low-confidence records: under 10\% deletion, the gap reaches
13.2 percentage points (95.58\% vs.\ 82.38\%), considerably larger than the
7.25-point gap observed under 20-meter positional noise, indicating
missing semantic context introduces greater inference uncertainty than
positional displacement alone.

\begin{table}[!htbp]
	\centering
	\caption{Prediction stability under random POI deletion.}
	\label{tab:poi_stability}
	\begin{tabular}{lccc}
		\hline
		\textbf{Deletion} & \textbf{All (\%)} & \textbf{High conf. (\%)} & \textbf{Low conf. (\%)} \\
		\hline
		5\%               & 93.28             & 96.25                    & 88.90                   \\
		10\%              & 92.43             & 95.58                    & 82.38                   \\
		\hline
	\end{tabular}
\end{table}

Table~\ref{tab:poi_per_activity} reveals a qualitatively different
activity-type pattern from the noise experiment. Home (99.72\%) remains
highly stable as its inference relies on spatial recurrence rather than POI
context. More notably, Work (93.29\%) and School (86.18\%) show considerably
greater sensitivity to POI deletion than to positional noise. Since Work and
School share similar temporal patterns, both concentrated during daytime
weekday hours, POI semantic information serves as the primary discriminating signal between them; its removal therefore degrades their
disambiguation rather than detection as mandatory activities. Among
non-mandatory activities, Leisure (75.91\%) and Shop Services (75.39\%) are
most affected, as these activities are characterized by functionally mixed
environments where any single POI carries high inferential weight.

\begin{table}[!htbp]
	\centering
	\caption{Per-activity prediction stability under 10\% POI deletion.}
	\label{tab:poi_per_activity}
	\begin{tabular}{cc|cc}
		\hline
		\textbf{Activity type} & \textbf{Ratio (\%)} & \textbf{Activity type}          & \textbf{Ratio (\%)} \\
		\hline
		1                      & 99.72               & 9                               & 75.91               \\
		2                      & 93.29               & 10                              & 82.37               \\
		3                      & 86.18               & 11                              & 87.94               \\
		4                      & 84.09               & 12                              & 79.02               \\
		5                      & 79.44               & 13                              & 80.10               \\
		6                      & 75.39               & 14                              & 79.56               \\
		7                      & 80.89               & 15                              & 86.01               \\
		8                      & 78.71               & \textbf{Activity-Weighted Avg.} & \textbf{92.43}      \\
		\hline
	\end{tabular}
\end{table}

Collectively, these experiments demonstrate that the proposed framework
achieves strong distributional alignment, produces well-calibrated confidence
scores, and generalizes across input degradation conditions.

\section{Conclusion and Future Work}
\label{sec:conclusion}

This paper presents a weakly supervised framework for trip purpose
inference, by assigning activity types to GPS-derived
staypoints without staypoint-level ground truth. The framework integrates
neighborhood-level POI semantic zones, differentiated inference strategies
for mandatory and non-mandatory activities, and confidence scoring calibrated
via multi-phase Pareto optimization against household survey reference
distributions. Experiments on over 81 million staypoints in Los Angeles
demonstrate consistent improvement over baseline across all distributional
alignment metrics, with a 48\% reduction in start time JSD.
Robustness analyses confirm that confidence scores reliably identify stable
predictions under GPS positional noise and incomplete POI coverage.

Despite its overall effectiveness, the framework is subject to two limitations. First,
although bias correction mechanisms are incorporated to mitigate GPS--survey discrepancies,
short-trip underreporting and recall errors in survey records remain partially
unresolved and warrant further investigation, as evidenced by residual divergence in early morning activity patterns. Additionally, commercial mobile-device
GPS data overrepresents younger, urban smartphone users and captures only periods
of active phone usage, potentially excluding rural and low-income populations and
introducing incomplete individual activity coverage.

Future work will explore learned GPS--survey reconciliation, sequential
activity context for spatially ambiguous non-mandatory types, and multi-regional survey calibration.

\section{Acknowledgement}
This work was supported in part by the FHWA Center for Excellence on New Mobility and Automated Vehicles Program, and in part by the Intelligence Advanced Research Projects Activity (IARPA) via Department of Interior/Interior Business Center (DOI/IBC) contract number 140D0423C0033. The U.S. Government is authorized to reproduce and distribute reprints for Governmental purposes notwithstanding any copyright annotation thereon. Disclaimer: The views and conclusions contained herein are those of the authors and should not be interpreted as necessarily representing the official policies or endorsements, either expressed or implied, of IARPA, DOI/IBC, or the U.S. Government.

\bibliographystyle{IEEEtran}
\bibliography{reference}

\end{document}